\documentclass[conference]{IEEEtran}
\IEEEoverridecommandlockouts
\usepackage{cite}
\usepackage{amsmath,amssymb,amsfonts}
\usepackage{algorithmic}
\usepackage{graphicx}
\usepackage{textcomp}
\usepackage{xcolor}
\usepackage{xcolor}

\usepackage{fancyhdr}

\fancypagestyle{firstpage}{%
    \lhead{2022 25th International Conference on Computer and Information Technology (ICCIT)
17-19 December, Cox’s Bazar, Bangladesh}

    \lfoot{979-8-3503-4602-2/22/\$31.00 ©2022 IEEE}
}

\def\BibTeX{{\rm B\kern-.05em{\sc i\kern-.025em b}\kern-.08em
    T\kern-.1667em\lower.7ex\hbox{E}\kern-.125emX}}

\makeatletter
\newcommand{\linebreakand}{%
  \end{@IEEEauthorhalign}
  \hfill\mbox{}\par
  \mbox{}\hfill\begin{@IEEEauthorhalign}
}
\makeatother

\begin{document}

\title{SleepExplain: Explainable Non-Rapid Eye Movement and Rapid Eye Movement Sleep Stage Classification from EEG Signal\\
{\footnotesize \textsuperscript{} }
\thanks{}
}

\author{\IEEEauthorblockN{Rafsan Jany}
\IEEEauthorblockA{
\textit{Islamic University of Technology}\\
 rafsanjany@iut-dhaka.edu}
\and
\IEEEauthorblockN{Md. Hamjajul Ashmafee}
\IEEEauthorblockA{
\textit{Islamic University of Technology}\\
ashmafee@iut-dhaka.edu}
\and
\IEEEauthorblockN{Iqram Hussain}
\IEEEauthorblockA{
\textit{Seoul National University}\\
iqram@ust.ac.kr}
\linebreakand 
\IEEEauthorblockN{Md Azam Hossain}
\IEEEauthorblockA{
\textit{Islamic University of Technology}\\
azam@iut-dhaka.edu}
%\and
%\IEEEauthorblockN{5\textsuperscript{th} Given Name Surname}
%\IEEEauthorblockA{\textit{dept. name of organization (of Aff.)} \\
%\textit{name of organization (of Aff.)}\\
%City, Country \\
%email address or ORCID}
%\and
%\IEEEauthorblockN{6\textsuperscript{th} Given Name Surname}
%\IEEEauthorblockA{\textit{dept. name of organization (of Aff.)} \\
%\textit{name of organization (of Aff.)}\\
%City, Country \\
%email address or ORCID}
}
\maketitle
\thispagestyle{firstpage}

\begin{abstract}
Classification of sleep stages is one of the most important diagnostic approaches for a variety of sleep-related disorders. Electroencephalography (EEG) is regarded as a powerful tool for examining the association between neurological effects and sleep phases since it correctly identifies sleep-related neurological alterations. During Non-Rapid Eye Movement (NREM) and Rapid Eye Movement (REM) sleep phases, a number of nerve and bodily functions are affected and therefore hold an important role both in their functionalities. This work aims to classify NREM and REM sleep stages from sleep EEG data and present a noble SleepExplain model, an explainable NREM and REM sleep stage classification to explain its predictions. In this work, sleep stages were classified using Random Forest, XGBoost, and Gradient Boosting ensemble classification models. Overall, we obtained an accuracy of 92.54\% (Random Forest), 94.25\% (Gradient Boosting), and 94.30\% (XGBoost). For explainable classification model, we utilized a game theoretic approach, SHAP (SHapley Addictive exPlanations) to offer a convincing explanation for the prediction.
\end{abstract}


\section{Introduction}
Sleep is one of the basic biological activities that are required for relieving stress. It is the brain's fundamental functions that are crucial for a person's learning ability, performance, and physical activity \cite{b2,b3}. Understanding sleep quality in an easier manner is the most important and interesting topic in the field of neuroscience and sleep disorder diagnosis. Sleep stage scoring is the state of the art for analyzing human sleep \cite{b4}. The goal of sleep stage scoring is to find the stages of sleep that are important for identifying and treating sleep disorders \cite{b5,b6}. 

The continuous recording of several electrophysiological signals, termed as Polysomnographic (PSG) signals, is used for sleep stage scoring purposes\cite{b7}. The American Academy of Sleep Medicine (AASM) provides the most widely used standard for sleep stage classification (AASM). According to this standard, PSG signal recordings are classified as Non-Rapid Eye Movement (NREM) sleep, Rapid Eye Movement (REM) sleep, and waking (W). The American Academy of Sleep Medicine (AASM) has more recent guidelines for this (AASM) \cite{b8,b9}. The AASM rules also specify the distinctive waves for each of the five sleep phases\cite{b12}. 

During NREM and REM stages, the human body has to face many functional changes in both the nervous system and the body system \cite{b18}. Hormonal changes also occur in these two stages \cite{b18}. When NREM sleep goes deep, sympathetic nerve functionality decreases. There is a break in sympathetic nerve activity at some point of NREM sleep due to the short increase in blood pressure and heart rate that follows K-complexes \cite{b18}. During REM sleep, respiratory flow and ventilation change and become faster and more erratic \cite{b19,b20}. Hypoventilation takes place in NREM sleep \cite{b21}. Significant reductions in blood flow and metabolism are linked to NREM sleep, while overall metabolic rate and blood flow during REM similar to wakefulness \cite{b22}.The alpha, beta, and gamma rhythms were attenuated in NREM sleep, while theta and delta rhythms rise with awakeness, followed by an increase in alpha and beta rhythms in REM sleep \cite{iqram}. Detecting and understanding NREM and REM sleep stages can provide many ways to detect nervous system and body functional disorders.

Except for the eye movement, middle ear ossicles, and respiratory system, the body is paralyzed during a Rapid Eye Movement (REM). Although the brain is less active during non-rapid eye movement (NREM), the body can still move. A sleep disorder like Narcolepsy is characterized by excessive daytime drowsiness and abnormal REM sleep regulation \cite{b23}. NREM sleep is related with the PARASOMNIAS or unusual sleep related behaviors, that take place while  sleeping \cite{b23}. Because the body muscles are more active. Predicting NREM and REM sleep is deemed to be a novel process for the diagnosis of sleeping disorders. It also evaluates  the body and nervous system characteristics during sleep. In the practical world, the sleep stage is performed through some manual processes. An expert measures and monitors the sleep scoring process manually. Notably, when it is a hand-operated procedure, errors can be coped up at any point. An automated strategy might be more reliable for regulating NREM and REM sleep.

Several EEG and biosignal studies have been published to investigate the relationship between EEG biomarkers and neurologic prognosis in medical and healthcare\cite{b34,b35,b36,b37,b38,b39}. The objective of this research is to develop a approach that can automatically classify NREM and REM sleep by explaining the prediction model using XAI. It will use multi-channel EEG signals to train machine learning models, and features will be retrieved from these signals. We intended to automate this sleep score technique using data from three EEG channels from three distinct sites (C4, O2, and F4). F4, C4, and O2 from the frontal, central, and occipital lobes were utilized for our investigation. The purpose of this study is to develop better models for forecasting NREM/REM sleep and explain those models using Explainable AI.

\section{Associated Study}
In the majority of research, overnight recorded EEG signals are used to classify the sleep stages \cite{b27}. These proposed studies recommend applying a variety of feature reduction strategies in order to identify the relevant features. Various classification methods have been developed to classify the phases of sleep. None of them, however, classifies sleep as exclusively NREM and REM sleep.

Satapathy et al.\cite{b24} proposed a method for identifying two stages of sleep, such as awake and sleep, using the ISRUC-Sleep\cite{b29} dataset. The overall accuracy for the awake and sleep stages was 91.67\% and 93.8\% respectively. Ellis et al.\cite{b25} provided an interpretable taxonomy of sleep stages, and their research classified sleep into five distinct phases. In this work, the PhysioNet Sleep-EDF\cite{b30} dataset was used. Santaji el al.\cite{b26} used another form of EEG data from sixty participants which were collected and preprocessed using an IIR (Infinite Impulse Response) in their research. In their study, three sleep classes (REM, NREM1, and NREM2) were classified with an overall accuracy of 95.36\%. Santosh et al.\cite{b27} presented a machine-learning model with an ensemble approach by using ISRUC-Sleep\cite{b29} dataset. In their work, sleep stages have been classified as waking, NREM (as N1, N2, and N3), and REM sleep achieving an accuracy 91.10\%. Shen et al.\cite{b28} proposed an enhanced machine learning model based on the essence of features applied to the ISRUC\cite{b30} dataset and classified sleep states into five stages, including waking, NREM (as N1, N2, and N3), and REM achieving an accuracy of 81.65\%. Hussain et al.\cite{iqram} used the neurological biomarkers of sleep phases to measure the delta wave power ratios (DAR, DTR, and DTABR). These measurements were evaluated by biomarkers due to their nature of decreasing during NREM sleep and increasing during REM sleep.

%In this section we are going to discuss the associated study regarding sleep stage classification. EEG signals are used in the majority of the studies. Most of the proposed experiments make use of EEG signals. These studies advise employing various feature reduction approaches in order to find the appropriate features. Finally, various classification models have been applied to categorize the sleep stages. But none of them classifies only NREM and REM sleep.

%In \cite{b24}, authors proposed using ISRUC-Sleep dataset for sleep stage scoring. They  classified two sleep stages, wake and sleep. The overall accuracy was 91.67\% and 93.8\%.  Ellis CA \cite{b25} proposed an explainable sleep stage classification. The PhysioNet Sleep-EDF dataset had been used in this study. Five sleep stages had been classified in this proposed study. In \cite{b26}, EEG data from 60 participants were captured and preprocessed using an IIR (Infinite Impulse Response) filter. Three class (REM, NREM1, NREM2) had been classified in this paper and the overall accuracy was 95.36\%. Santosh  \cite{b27} proposed machine learning with ensemble staking model by using ISRUC-Sleep datase. Five classes had been classified in this study. 91.10\% accuracy was achieved. In \cite{b28}, authors proposed an improved model based essence features applied on ISRUC dataset. The accuracy was 81.65\% for five classes. 

\section{Material And Methodology}
This study presents an efficient and reliable automatic sleep stage classification model of NREM and REM sleep classes on three-channel EEG signals.  Alongside, we added an explainable AI (XAI) appraoch for better understanding the inside mechanism of the model. The explainable AI shows the most devoted features for the final outcome produced by the classification model. Notably, we focused on increasing the accuracy of the prediction model of the NREM and REM sleep stages and explain this model with SHAP based approach.  

\subsection{Model Architecture}
\begin{figure}[h]
    \hspace*{-0.4cm}
    %\label{method}
     \includegraphics[width=1\linewidth,height=0.65\linewidth,clip, trim = 0 0 0 0,]{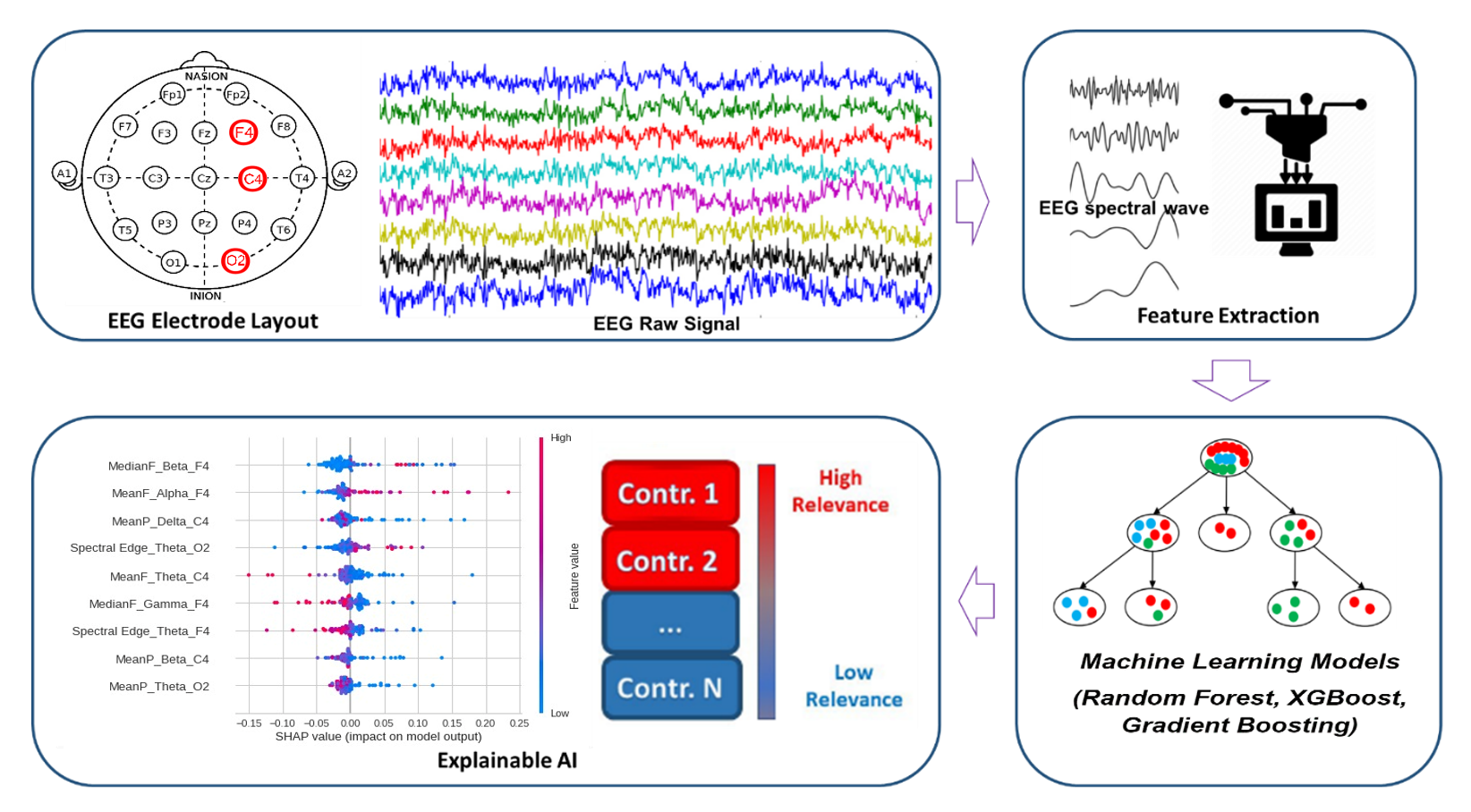}
    \caption{Proposed framework of an explainable sleep stage classification model.}
    \label{fig}
\end{figure}

Figure \ref{fig} depicts a general architecture of the proposed investigation. First, we acquired the EEG data from channels F4, C4, and O2 and extracted features from the raw signal using FFT (Fast Fourier Transform). To eliminate any 60 Hz AC noise from the neighboring electrical grid, the EEG signal was filtered. Then, from the noise-free signal, features are retrieved. Next, the model was trained using three ensemble machine learning models: Random Forest, Gradient Boosting, and XGBoost. Finally, we implemented an SHAP based explainable classification model to explain the outcomes produced by the prior model.

\subsection{Dataset Description}
The dataset was collected from a sleep center named Haaglanden Medisch Centrum (HMC, The Netherlands)\cite{b32,b33}. The initial data files are combined with different types of signals. We selected EEG and F4 from frontal lob, C4 from central lob, and O2 from occipital lob. Frontal lob controls the voluntary movements of the body. The thoughts and analytical activities are controlled from central lob. Occipital lob is responsible for the eyesight. To characterize the sleep stages of a cerebral condition, the EEG is divided into five frequency subbands: delta wave (0Hz - 4Hz), theta wave (4Hz - 8Hz), alpha wave (8Hz - 12Hz), beta wave (12Hz - 30Hz), and gamma wave (30Hz). Our dataset contains 154 sleep recordings with 75 features. There are five classifications in the dataset, such as Wake, N1, N2, N3, REM. In the preprocessing stap, all wake classified rows were removed and N1, N2, N3 classes were merged into a single NREM class. So we have received the final preprocessed dataset having two classes; NREM and REM. The final dataset has a total of 89096 rows, where rows containing NREM are 72631 and rows containing REM are 16465. The difference between the sample counts of NREM and REM classes is significant. So we implemented the SMOTE (Synthetic Minority Over-sampling Technique)\cite{b13} approach to balance our training dataset. The testing dataset was unchanged to find the original accuracy of our proposed model.

\subsection{Classification Models}
We used three popular machine learning models Random Forest, Gradient Boosting and, XGBoost. The training dataset was balanced using SMOTE technique. The default values of performance parameters like \emph{n\_estimators} and \emph{max\_depth} did not produce a satisfactory result. So we had to tune the parameters of the algorithms. We used scikit-learn python library for training and tuning our models.

\begin{table}[h]
\caption{Optimum Parameter Values} % title of Table
\centering % used for centering table
\begin{tabular}{c c c c c} % centered columns (4 columns)
\hline\hline %inserts double horizontal lines
\ Parameters\ & RF\ & GB\ & XG\\ 
[0.5ex] 
\hline 
max\_depth & 39 &   12 & 29  \\
n\_estimators & 450 & 1150 & 4010  \\
 %\\[1ex] % [1ex] adds vertical space
\hline %inserts single line
\end{tabular}
\label{table:Per_Mat} % is used to refer this table in the text
\end{table}

Random forest is one of the most effective machine learning techniques for classification\cite{b14}. In this solution, we tuned this model with the \emph{n\_estimators} covering the range of 3 to 500 and the \emph{max\_depth} covering the range of 5 to 50. We figured out the best result when, \emph{n\_estimators} = 450 and \emph{max\_depth} = 39 having the accuracy of 92.52\%

Using an iterative process, boosting algorithms merge weak learners into a strong learner\cite{b15}. Gradient boosting is a regression approach that resembles boosting\cite{b16}. The accuracy with the default value of \emph{n\_estimators} and \emph{best\_dept}
for Gradient Boosting Classifier was not quite acceptable. We tuned this model at the parameters, \emph{n\_estimators} having the range of 5 to 1200 with a interval of 50 and \emph{best\_dept} having the range of 3 to 30. The most acceptable result was acquired for \emph{n\_estimators} = 1150 and \emph{best\_dept} = 12. The accuracy with this tuning is 92.25\%.

Another ensemble model based on Gradient Boosting with a high degree of scalability is called XGBoost\cite{b17}. XGBoost constructs a loss function which is minimized if the objective function is expanded additively, similar to Gradient Boosting. The accuracy with the default value of \emph{n\_estimators} and \emph{best\_dept} for XGBClassifier Classifier was not also satisfactory. This model was tuned with the parameters, \emph{n\_estimators} including the range of 500 to 5000 with an interval of 50 and the \emph{best\_dept} including the range of 3 to 30. The topmost result of this model was attained for \emph{n\_estimators} = 4010 and \emph{best\_dept} = 29. It scored 94.3\%.

\subsection{Explainability of the Proposed Model using SHAP}
It is crucial to be able to appropriately interpret the results of prediction models. It fosters the right level of user trust, offers suggestions for how to make a model better, and aids in comprehending the learning process that is being represented\cite{b40}. Explainable artificial intelligence (XAI) helps users to understand and believe the result produced by machine learning algorithms. Each feature is given a relevance value by SHAP (SHapley Additive exPlanations) for a specific prediction\cite{b1}. In this study, we took out the most important features and their contributions for a particular prediction applying SHAP value. 

It explains an outcome of a model by computing the contribution of each feature associated with it. The Shapely value of an outcome is produced following a linear model and measures how much each feature in that model contributes, either positively or negatively. 

% It specifies the explanation for an instance, \textit{x} as:

% % %%%
% % Add equation here
% % also complete the items mentioned below from the paper directly: 
% % ok sir, but x or z ? line 114...ok sir got it
% % %%%
% \begin{equation}
% g(z') = \Phi_0 + \[ \sum_{j=1}^{M} \Phi_j z'_j \]
% \end{equation}

% where:
% \begin{itemize}
%   \item \textit{g} is the explanation model.
%   \item \textit{z'} ple
% is the coalition vector (also called simplified features),
% and z
% 0 ∈ {0, 1}^M $$
%   \item \textit{M} is the maximum coalition size. 
%   \item \textit $\Phi_j$ belongs to real number  is the feature attribution for the feature j for instance x. It is the Shapley value. 

% \end{itemize}

\section{Result And Discussion}
In this section we will discuss the outcome of prediction performances. LaterSHAP detects the most influential features for XGBoost classifier. 

\begin{table}[h]
\caption{Performance Metric Results For Models} % title of Table
\centering % used for centering table
\begin{tabular}{c c c c c} % centered columns (4 columns)
\hline\hline %inserts double horizontal lines
\ Evaluation Metric\ & RF\ & GB\ & XG\\ 
[0.5ex] 
\hline 
Accuracy & 92.54\% &   94.25\% & {\bf94.30}\%  \\
Precision & 87.13\% & 91.25\% & {\bf91.3}3\%  \\
Recall  & 89.31\% & 89.65\% & {\bf89.70}\% \\
Specificity  &  89.31\%  & 89.51\% & {\bf89.70}\%  \\ 
F1 Score  &  88.21\%  &  90.44\% & {\bf90.51}\%   \\[1ex] % [1ex] adds vertical space
\hline %inserts single line
\end{tabular}
\label{table:Per_Mat} % is used to refer this table in the text
\end{table}

\subsection{Model Performance Result}

In Table \ref{table:Per_Mat}, the performance metric of three models are given. The best accuracy was achieved by XGBoost classifier and it was 94.30\%, when other two classifiers Random Forest and Gradient Boost scored 92.54\% and 94.25\% respectively. XGBoost and Gradient Boosting scored almost same score. The differences is very tiny.  In other Evaluation metrics, XGBoost performed very satisfactory. Gradient Boosting was also little bit behind from XGBoost. But Random Forest was far behind from other two classifiers. Individually, 92.54\% is not a bad accuracy. But compare to other two classifiers, Random Forest was far behind. 

\begin{figure}[h]
    \hspace*{-0.3cm}
     \includegraphics[width=1.2\linewidth,height=0.9\linewidth,clip, trim = 110 0 0 30,]{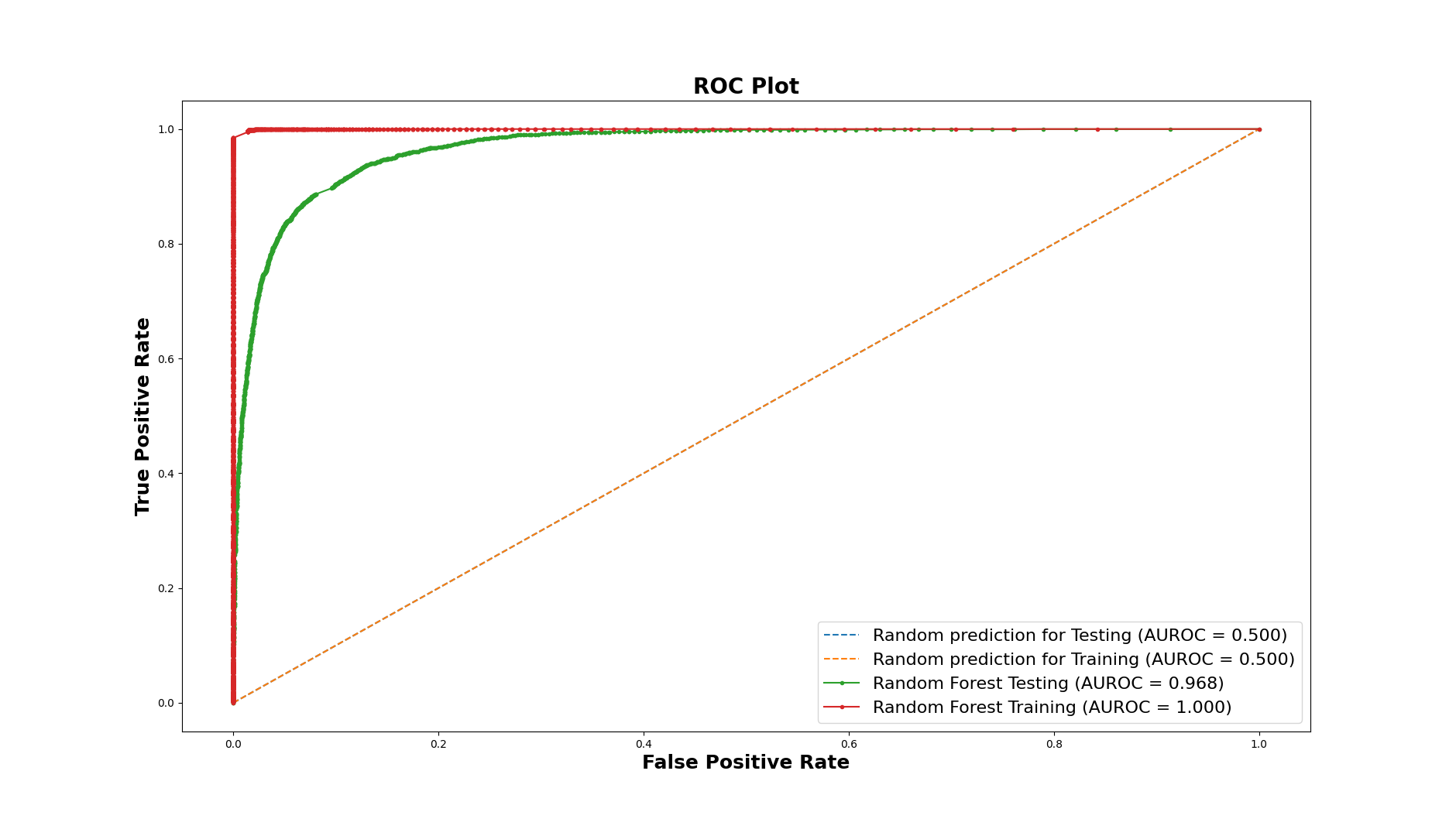}
    \caption{ROC Graph of Random Forest Model}
    \label{fig:RF_ROC}
\end{figure}

The ROC curve illustrates the trade-off between specificity and sensitivity. A better performance is shown by classifiers that provide curves that are closer to the top-left corner. The figure \ref{fig:RF_ROC} is presenting the training-testing ROC graphs of Random Forest. In the top-left corner, differences between training and testing are slightly large than other two models. The AUROC for the training is 1.00, and for the testing the AUROC is 0.968.

\begin{figure}[h]
   \hspace*{-1.4cm}
     \includegraphics[width=1.3\linewidth,height=0.9\linewidth,clip, trim = 0 0 0 0,]{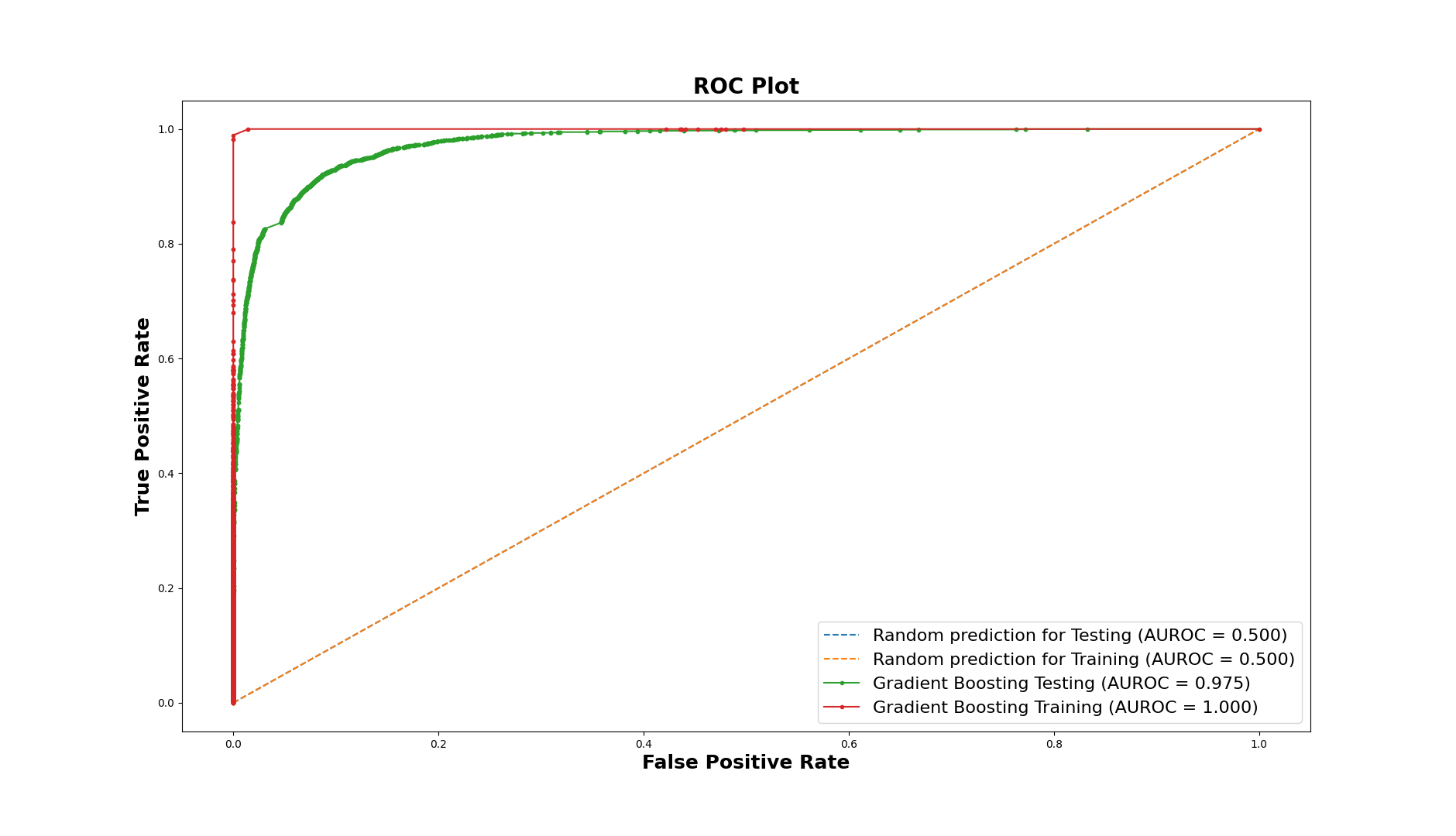}
    \caption{ROC Graph of Gradient Boosting Model}
    \label{fig:GB_ROC}
\end{figure}

The training-testing ROC graph of Gradient Boosting 
is shown in The figure \ref{fig:GB_ROC}. The differences of the top-right corner of the graph was reduced The training and testing curves are closer than the Random Forest model. The accuracy is better  The AUROC is 0.975 for the testing and 1.00 for the training.

\begin{figure}[h]
    \hspace*{-1.4cm}
     \includegraphics[width=1.3\linewidth,height=0.9\linewidth,clip, trim = 0 0 0 0,]{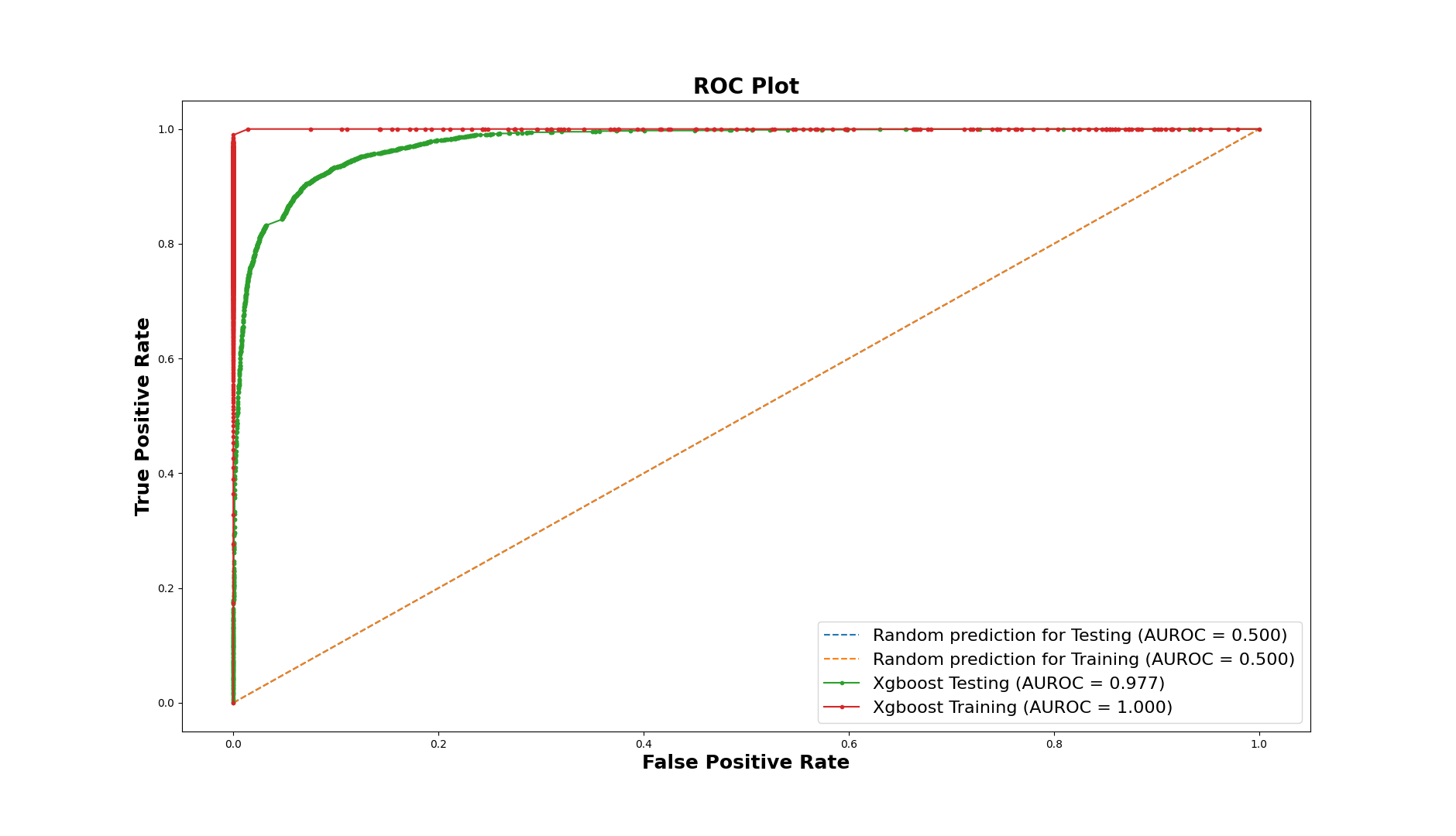}
    \caption{ROC of XGBoost Model}
    \label{fig:XG_ROC}
\end{figure}

The figure \ref{fig:XG_ROC} is showing the training-testing ROC graphs of XGBoost. XGBoost performed better than other two models. The differences between XGBoost and Gradient Boosting is not not much. That's why, the ROC curvs of this two models are significantly similar. The testing curve is more closer to the training graph than the other two graphs. The AUROC is 0.977 which close to the Gradient Boosting  \ref{fig:GB_ROC}. There are no unusual ups and down in the ROC graphs of the three models. The Training and testing graphs were naturally smooth. There are no unusual ups and down in the ROC graphs of the three models. The Training and testing graphs were naturally smooth. Though, we used SMOTE for the balancing the training data, the models performed well with the uncorrupted testing data. So we can assure that, the models are not overfitted.

\begin{figure}[h]
    \hspace*{+.8cm}
     \includegraphics[width=0.6\linewidth,height=0.5\linewidth,clip, trim = 0 0 0 0,]{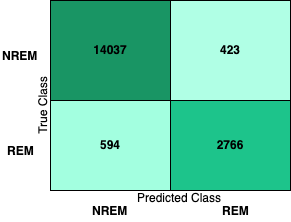}
    \caption{Confusion Matrix of XGBoost Model}
    \label{fig:XG_CON}
\end{figure}

Classification problem's prediction outcomes are compiled in a confusion matrix. In the figure \ref{fig:XG_CON}, the confusion matrix of XGBoost is shown. We can see the total prediction of NREM classes were 14460, where 14037 predictions were correct and only 423 predictions were wrong. Total REM classes wee 3350, 2766 were correct and 584 were wrong. We did not use SMOTE in testing data. That's why, the imbalance is noticeable here. Though our models did perform well.

\subsection{SHAP(XAI) Result}
The internal mechanism of a machine learning model is hard to understand and difficult to relate to the practical dataset and scenario. It is deemed to be a novel approach to use Explainable AI(XAI) to understand the inner mechanism of a model. In this study, we used SHAP for the better understanding and feature dependencies. We applied SHAP in XGBoost classifier to determine the most significant features for a particular prediction. We considered the first 100 rows of the testing data.  

\begin{figure}[h]
    \hspace*{+.7cm}
     \includegraphics[width=0.8\linewidth,height=0.5\linewidth,clip, trim = 0 0 0 0,]{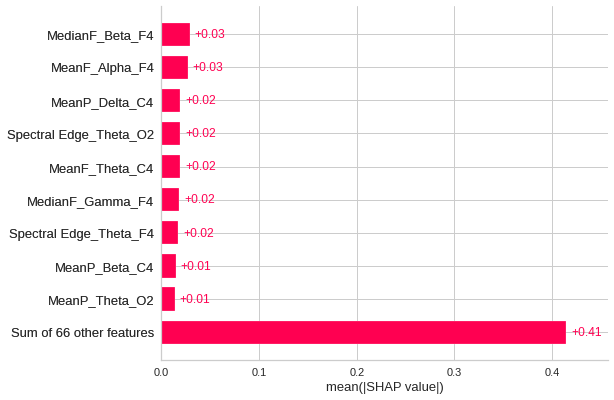}
    \caption{Significant Features With SHAP Value}
    \label{fig:shap_bar}
\end{figure}

Fig \ref{fig:shap_bar} is showing the list of the most important features and their SHAP values for predictions. MedianF\_Beta\_F4 is the most dedicated feature. MeanF\_Alpha\_F4 is also scored almost same as Median\_Beta\_F4. Most of the features of the list scored +0.02. The rest of the features scored +0.41 altogether. Notably, the most of the significant features are from the frontal lob. The dedication of the occipital lob is less than other two lobs.

\begin{figure}[h]
    \hspace*{+.7cm}
     \includegraphics[width=0.8\linewidth,height=0.5\linewidth,clip, trim = 0 0 0 0,]{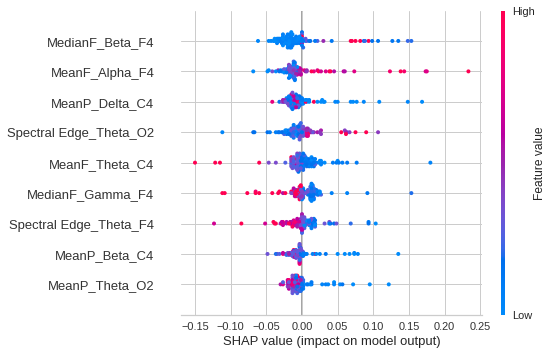}
    \caption{SHAP Value Beeswarm}
    \label{fig:shap_beeswarm}
\end{figure}

The beeswarm plot(fig \ref{fig:shap_beeswarm}) is made to show a summary of the top features in a dataset and how they affect the model's output in a way that is both information-dense and easy to understand. A single dot is used to indicate each instance of the explanation in each aspect of the figure \ref{fig:shap_beeswarm}. Here, MedianF\_Beta\_F4 is the most important feature on average. So from the both figure (fig \ref{fig:shap_bar} and fig \ref{fig:shap_beeswarm}), we can notice one thing and that is , F4 and C4 have a very significant role, where O2 has not such kind of contribution like other two lobes. The duration of REM sleep was less then the NREM. Eyesight and eye movement are controlled by the occipital lob. This is cause to reduce the supremacy of O2 lob in the model.

\section{Conclusion}

In this study, NREM and REM sleep phases were classified using ensemble classification models from Random Forest, XGBoost, and Gradient Boosting. Overall, Random Forest obtained 92.54\% accuracy, XGBoost earned 94.30\% accuracy, and Gradient Boosting also achieved 94.25\% accuracy. We also applied SHAP in the XGBoost classifier to determine the most significant features for a particular prediction. Most of the traditional classifying approaches are for the five stage classification of sleep. Very few studies worked with XAI. So the inside mechanism of a model is not revealed. It remained a black box model. This study unleashed the model explainability with XAI and classified the two most significant sleep stages (NREM and REM) with better accuracy.

\section*{Acknowledgment}
This research was supported by Islamic University of Technology Research Seed Grants (IUT RSG) (Ref: REASP/IUT-RSG/2022/OL/07/012).

\vspace{12pt}
% there is an error in equation i can not solve it

\end{document}